\documentclass[letterpaper, 10 pt, conference]{ieeeconf}  %

\IEEEoverridecommandlockouts                              %

\let\labelindent\relax
\usepackage{amsmath}
\usepackage{amssymb}
\usepackage{esvect}
\usepackage{tabularx}
\usepackage{booktabs}
\usepackage{enumerate}
\usepackage{array,multirow}
\usepackage{tablefootnote}
\usepackage{enumitem}
\usepackage{caption}
\usepackage{subcaption}
\usepackage[export]{adjustbox}
\usepackage[table]{xcolor}
\usepackage{pifont}
\usepackage[T1]{fontenc}
\usepackage{soul} %

\usepackage{flushend}

\makeatletter
\let\NAT@parse\undefined
\makeatother
\usepackage[pagebackref=true,breaklinks=true,letterpaper=true,hidelinks,bookmarks=false]{hyperref}
\newcommand{\cmark}{\ding{51}}%
\newcommand{\xmark}{\ding{55}}%

\definecolor{mylocerror}{RGB}{213, 94, 0}
\definecolor{myfn}{RGB}{0, 114, 178}
\definecolor{myfp}{RGB}{0, 158, 115}
\definecolor{myids}{RGB}{204, 121, 167}
\definecolor{myGTann}{RGB}{0, 160, 0}
\definecolor{myforecast}{RGB}{255, 165, 0}
\definecolor{myegocar}{RGB}{200, 0, 0}
\definecolor{myroadline}{RGB}{0, 190, 255}
\definecolor{mylaneboundary}{RGB}{173, 0, 255}
\definecolor{mydrivablearea}{RGB}{0, 0, 255}
\definecolor{staticcars}{RGB}{173, 173, 173}

\usepackage{soul}
\usepackage[T1]{fontenc}  
\usepackage[cmintegrals]{newtxmath}
\usepackage{bm}
\usepackage{amsmath}

\usepackage{xspace} 
\usepackage{multirow} 

\usepackage{graphicx}
\graphicspath{{figures/}}

\usepackage{relsize}
\usepackage{nicematrix}

\usepackage{array}

\usepackage{makecell}

\usepackage{url}

\newcommand{\awta}{aWTA\xspace}

\definecolor{exemplargreen}{HTML}{f3d6d7}
\definecolor{searchred}{HTML}{f3d6d7}
\definecolor{darkgreen}{HTML}{445b00}
\definecolor{lightergray}{HTML}{dddddd}
\definecolor{goodgreen}{rgb}{0.0, 0.56, 0.0}
\definecolor{badgray}{HTML}{666666}

\definecolor{buscasota}{rgb}{0.9, 0.95, 1.}

\newcommand{\gooddelta}[1]{\scalebox{0.9}{($\bm{\textcolor{goodgreen}{#1}}$)}}

\definecolor{rankacolor}{HTML}{bebeff}
\definecolor{rankbcolor}{HTML}{d9d9ff}
\definecolor{rankccolor}{HTML}{e9e9ff}

\newcommand{\plusours}{$w/$ \textbf{aWTA\xspace}}

\title{\LARGE \bf
Annealed Winner-Takes-All for Motion Forecasting
}

\author{Yihong Xu$^{1^*}$ Victor Letzelter$^{1,2^*}$  Micka\"el Chen$^1$ \'Eloi Zablocki$^1$ Matthieu Cord$^{1,3}$
\thanks{$^{1}$ Valeo.ai, Paris, France; Email: \texttt{firstname.lastname@valeo.com}}
\thanks{$^{2}$ LTCI, T\'el\'ecom Paris, Institut Polytechnique de Paris, France.}
\thanks{$^{3}$ Sorbonne Universit\'e, Paris, France.}%
\thanks{$^*$ Equal contribution.}
\thanks{Corresponding author: Y. Xu, \texttt{yihong.xu@valeo.com}}
} 

\begin{document}

\maketitle
\thispagestyle{plain}
\pagestyle{plain}

\begin{abstract}
In autonomous driving, motion prediction aims at forecasting the future trajectories of nearby agents, helping the ego vehicle to anticipate behaviors and drive safely.
A key challenge is generating a \emph{diverse} set of future predictions, commonly addressed using data-driven models with Multiple Choice Learning (MCL) architectures and Winner-Takes-All (WTA) training objectives.
However, these methods face initialization sensitivity and training instabilities.
Additionally, to compensate for limited performance, some approaches rely on training with a large set of hypotheses, requiring a post-selection step during inference to significantly reduce the number of predictions.
To tackle these issues, we take inspiration from annealed MCL, a recently introduced technique that improves the convergence properties of MCL methods through an annealed Winner-Takes-All loss (\awta).
In this paper, we demonstrate how the \awta loss can be integrated with state-of-the-art motion forecasting models to enhance their performance using only a minimal set of hypotheses, eliminating the need for the cumbersome post-selection step. Our approach can be easily incorporated into any trajectory prediction model normally trained using WTA and yields significant improvements. To facilitate the application of our approach to future motion forecasting models, the code is made publicly available: \url{https://github.com/valeoai/MF_aWTA}.
\end{abstract}

\section{Introduction}
\label{sec:intro}

Motion forecasting is essential for the safe deployment of autonomous driving systems as it allows vehicles to anticipate the possible behaviors of nearby agents and plan their actions accordingly.
A key requirement for motion forecasting is that models should propose multiple plausible futures for each agent.
This diversity is essential for downstream tasks since anticipating even rare cases is crucial for the safety of autonomous agents \cite{mfp,park2020diverse,cui2021lookout,calem2022diva}.
However, for any given situation, only one realization of a future trajectory is observed in the data.
This poses a challenge because the vast majority of current methods are data-driven \cite{salzmann2020trajectron++,kim2021lapred,mtr,cui2023gorela,xu2024towards,xu2024matrix,ettinger2024scaling,feng2024unitraj}.

\begin{figure}[ht]
\centering
\begin{subfigure}{0.32\linewidth}
\includegraphics[trim={3cm 0 0 0},clip,width=\linewidth]{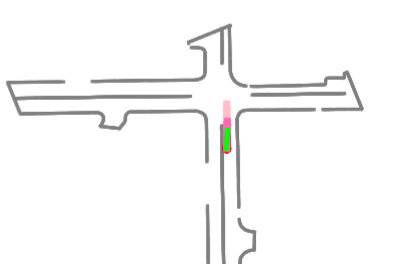} 
  \caption{WTA 6 hyp.}
  \label{fig:wta_6}
\end{subfigure}
\hfill
\begin{subfigure}{0.32\linewidth}
\includegraphics[trim={3cm 0 0 0},clip,width=\linewidth]{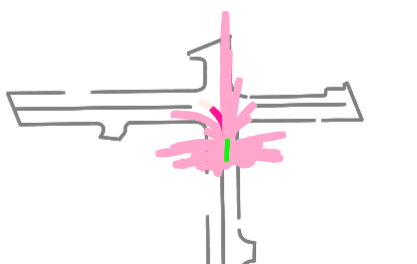}
  \caption{WTA 64 hyp.}
  \label{fig:wta_64}
\end{subfigure}
\hfill
\begin{subfigure}{0.33\linewidth}
\includegraphics[trim={3cm 0 0 0.1cm},clip,width=\linewidth]{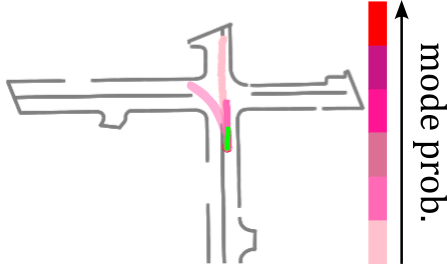}
  \caption{\textbf{\awta} 6 hyp.}
  \label{fig:awta_6}
\end{subfigure}
\caption{\textbf{Issues of WTA, and our proposed \awta.} Motion forecasting models trained with WTA and a small number of hypotheses (e.g., 6 hyp.) suffer from mode collapse (a), causing a performance drop. A naive solution to improve performance is to increase the training  hypotheses (b), but a cumbersome post-selection is required. Alternatively, \awta (c) covers more effective modes while consistently using the same minimum number of hypotheses for both training and inference, discarding the post-selection and achieving better performance, for example, -9.41\% minADE and -36.67\% MissRate with MTR~\cite{mtr} versus (b) after selection.
}\label{fig:teaser}
\end{figure}

The challenge of training a multi-hypothesis model with only single observations is commonly addressed using Multiple Choice Learning (MCL) with a Winner-Takes-All (WTA) loss \cite{guzman2012multiple, lee2016stochastic}.
In this framework, multiple hypotheses --- corresponding to possible futures (modes) --- are predicted and scored by the model.
With WTA training, only the best-performing 
prediction head (or \textit{hypothesis}) is selected and updated during training.
However, as an example shown in~\autoref{fig:teaser}, the naive use of WTA can lead to instabilities, such as mode collapse during training \cite{RWTA}.
To mitigate this, many motion forecasting models increase the number of proposals \cite{Wayformer} or use `intention points' \cite{mtr} to compensate for the performance drop from mode collapsing.
However, to be effective, these approaches generally use a very large number of proposals.
To reduce the number of predictions at inference time, a selection method like non-maximum suppression (NMS) or clustering is often applied~\cite{mtr, Wayformer, mtreda}.
These strategies make model performance highly sensitive to the proposal selection method and are computationally inefficient.
While this issue has been explored in other fields, leading to several proposed solutions \cite{RWTA, ilg2018uncertainty, firman2018diversenet}, these findings have been scarcely applied to autonomous driving problems \cite{dac, EWTA}.

In this work, we take inspiration from a recently introduced training scheme for MCL, called \emph{annealed} Multiple Choice Learning \cite{amcl}.
This method employs the `annealed WTA' (\awta) loss, 
a theoretically grounded approach that has shown promising results primarily on synthetic or toy datasets and small-scale models.
Motivated by its potential, we propose to adopt the \awta loss for motion forecasting models.
In contrast to the simpler use cases explored in the original work \cite{amcl}, we demonstrate that this optimization scheme is particularly well suited for the trajectory prediction task.
We show that this learning scheme can be effectively integrated into larger models, real-world experiments, and large-scale datasets.
We find that \awta systematically improves over current alternatives \cite{dac,RWTA,EWTA} across various recent forecasting models~\cite{mtr, Wayformer} and two large-scale real-world datasets~\cite{argoverse2,wmod}.
Furthermore, the use of \awta drastically reduces the necessary number of hypotheses, eliminating the need for cumbersome test-time selection methods.
Importantly, our proposal can be easily incorporated into any trajectory prediction model that normally uses WTA, which includes the vast majority of modern motion forecasting models.

\section{Related Work}

Motion forecasting consists of predicting the future trajectory $y \in \mathcal{Y}$ for an agent, e.g., an autonomous vehicle, given a road map and a history of states as grasped in $x \in \mathcal{X}$. 
Current motion forecasting methods~\cite{salzmann2020trajectron++, kim2021lapred, mtr, cui2023gorela, Wayformer, covernet, yuan2021agentformer, densetnt, multipath, multipath++, mfp, mtp, mosa, wang2023ganet, xu2024valeo4cast, xu2024ppt} focus on designing architectures to better take into account the map information surrounding the agent (i.e., vehicle to forecast), the behaviors of neighboring agents, and the history of their own trajectories.
For example, LaPred \cite{kim2021lapred} enforces the use of map information when predicting the future trajectories of agents of interest. This is achieved by integrating the closest vectorized lane information and the past trajectories of neighboring agents into the agent features.
More recent methods are primarily based on transformers. For instance, Wayformer~\cite{Wayformer} stacks transformer attention modules to encode the road graph, the agent history, and the agent interaction. Different modalities are then aggregated by a scene encoder, and the merged features are used for trajectory decoding, with randomly initialized and learnable mode queries and fully-connected layers.
MTR \cite{mtr} follows a similar transformer-based structure but introduces `intention points' that provide prior information on the future endpoints to initialize their queries.

Different methods to learn future trajectories have been proposed. In particular, goal-based methods~\cite{zhao2021tnt} define sparse anchors and learn to predict their confidence scores. Based on these scores, some anchors are selected as goals, and the method completes the trajectory for each goal. 
Heatmap-based methods~\cite{gilles2021home} follow a similar idea but predict a heatmap with score responses that indicate the goals. However, most current state-of-the-art methods still follow the more straightforward approach of regression, which directly learns to predict plausible future trajectories using the commonly used Winner-Takes-All (WTA) loss. These methods achieve better performance compared to other heuristic approaches~\cite{gu2023vip3d, uniad}.

Yet, while the WTA loss is known to be unstable during training \cite{RWTA}, only a handful of works have examined this issue in motion forecasting pipelines \cite{RWTA, EWTA, dac}. For example, $\bullet$ \textbf{Relaxed WTA (RWTA)} \cite{RWTA} which updates both the winning head and the non-winning ones, with the latter being scaled by a small constant $\varepsilon=0.05$, $\bullet$ \textbf{Evolving WTA (EWTA)} \cite{EWTA} which updates the top-$n$ heads instead of just the winner, with $n$ following a 
decreasing schedule during
training, $\bullet$ \textbf{Divide and conquer (DAC)}~\cite{dac} which recursively splits the set of hypotheses indices into two subsets until a certain depth, which depends on the training step. 
These WTA variants can be unified under the common formulation 
with a hypotheses-weighted loss (described in the following section)
where the scheduled weight $q_t$ depends on the specific method, as detailed in Table \ref{tab:losses}.
\begin{table}[t]
\centering
\caption{
\textbf{Expression of $q_t(f_k, x, y)$ in WTA variants, with $\mathcal{L}_t$ from \eqref{eq:loss_expr}}. Here, $\mathbf{1}_{\mathrm{WTA}}(k) \triangleq \mathbf{1}[k \in \mathrm{argmin}_s \ell(f_s(x),y)]$, where $k^{\star} = \mathrm{argmin}_s \ell(f_s(x),y)$. $\mathcal{T}_{n}(\theta,x,y)$ refers to the best $n$ hypotheses for $(x,y)$, while $\mathcal{P}(d_t)$ is the hypotheses set at depth $d_t$ (as described in \cite{dac}). $Z^{-1}$ ensures the weights normalization when needed.
The column 
`Schedule' (`Sched.') indicates if the weight $q_t$ depends on the step $t$. aWTA is unique in having an infinite, uncountable set $\mathcal{V}$ for $q_t$ values, offering more flexibility.
}
\resizebox{1.0\columnwidth}{!}{
\begin{tabular}{@{}l @{\hspace{0.2cm}} c @{\hspace{0.1cm}} c @{\hspace{0.1cm}} c@{}}
    \toprule
    Method & $q_t(f^k_{\theta}, x, y)$ & Sched. & Values $\mathcal{V}$\\
    \midrule
    WTA & $\mathbf{1}_{\mathrm{WTA}}(k)$ & \xmark & $\{0,1\}$\\
    RWTA \cite{RWTA} & $(1-\frac{K}{K-1} \varepsilon) \mathbf{1}_{\mathrm{WTA}}(k) + \frac{\varepsilon}{K-1}$ & \xmark & $\{1-\varepsilon, \varepsilon/(K-1)\}$\\
    EWTA \cite{EWTA} & $Z^{-1} \mathbf{1}[k \in \mathcal{T}_{n(t)}(\theta,x,y)]$ & \cmark & $\{0, Z^{-1} \}$\\
    DAC \cite{dac} & $ Z^{-1} \sum_{\mathcal{S} \in \mathcal{P}(d_t) \;|\;k \in \mathcal{S}} \mathbf{1}[k^{\star} \in \mathcal{S}]$ & \cmark & $|\mathcal{V}|<\infty$, $\mathcal{V} \subset \mathbb{Q}_{+}$\\
    \textbf{\awta} \cite{amcl} & $Z^{-1} \exp \left(-\ell\left(f_k(x), y\right) / T(t)\right)$ & \cmark & $|\mathcal{V}| = \infty$, $\mathcal{V} \subset \mathbb{R}_{+}$\\
    \bottomrule
\end{tabular}
 or}
\label{tab:losses}
\end{table}

Unlike previous works that are limited to small models and mostly simulated trajectories, our work 
proposes for motion forecasting an alternative called annealed Winner-Takes-All (aWTA).
Our approach outperforms WTA and its variants \cite{RWTA, EWTA, dac} in large-scale datasets with modern motion forecasting models.

\section{Annealed Winner Takes All (aWTA) in Motion Forecasting}

\subsection{WTA for Motion Forecasting}
\label{sec:wta_losses}

Current regression-based motion forecasting models leverage WTA as the training objective to generate $K$ diverse and plausible future trajectories. Given a past trajectory and map information $x$, a motion forecasting model parameterized by $\theta$ with $K$ forecasting heads produces $K$ future trajectory predictions $f^1_{\theta}, \dots, f^K_{\theta}$.

During training, knowing its corresponding future ground-truth trajectory $y$, the WTA loss only optimizes the head that produces the prediction with the minimum cost $\ell$:
\begin{align}
    k^{\star} = \mathrm{argmin}_k \ell(f^k_{\theta}(x),y),
    \label{eq:argmin}
\end{align} 

 The $k^{\star}$-th head is then trained to minimize: 
 \begin{align}
      \mathcal{L}(\theta) = \ell(f^{k^{\star}}_{\theta}(x),y),
      \label{eq_wta}
 \end{align}
Typically, $\ell$ is the Average (euclidean) Distance Error (ADE) \cite{salzmann2020trajectron++,benyounes2022cab}, 
 \begin{align}
     \ell\left(f^k_{\theta}(x), y\right) = \frac{1}{L}\sum_{j=1}^{L}(f^k_{\theta}(x)_j- y_j)^2,
 \end{align} 
where $L$ {is the number of steps of the} future trajectory. Alternatively, a Gaussian negative log-likelihood loss is also often used~\cite{salzmann2020trajectron++, mtr, Wayformer}, considering each prediction as the component of a Gaussian mixture, which remains applicable with WTA and \awta. 
\begin{figure*}[ht]
\centering

\begin{subfigure}{0.49\linewidth}
\includegraphics[trim={0 0 0 0},clip,width=\linewidth]{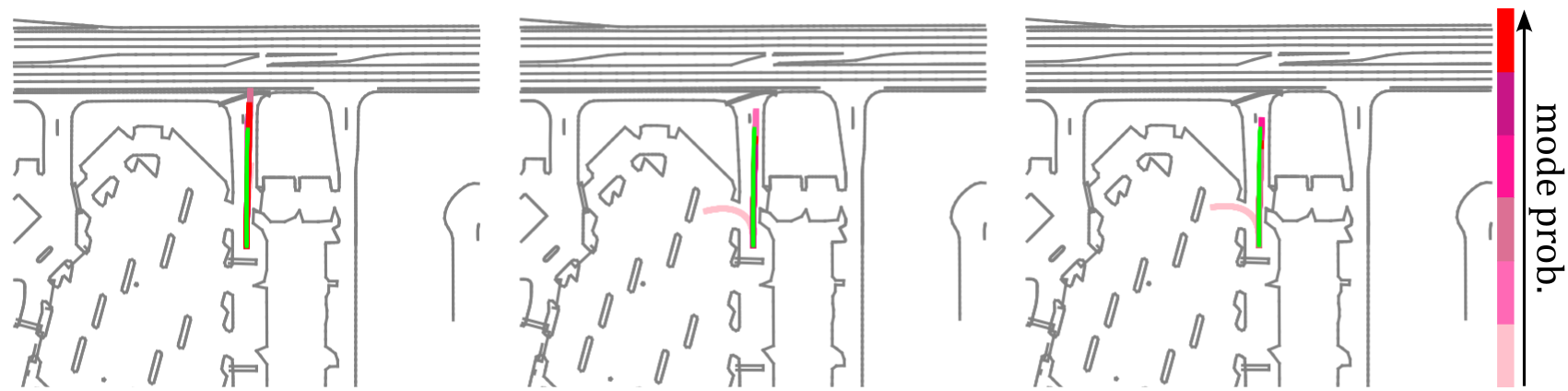} 
\end{subfigure}
\begin{subfigure}{0.49\linewidth}
\includegraphics[trim={0 0cm 0 0},clip,width=\linewidth]{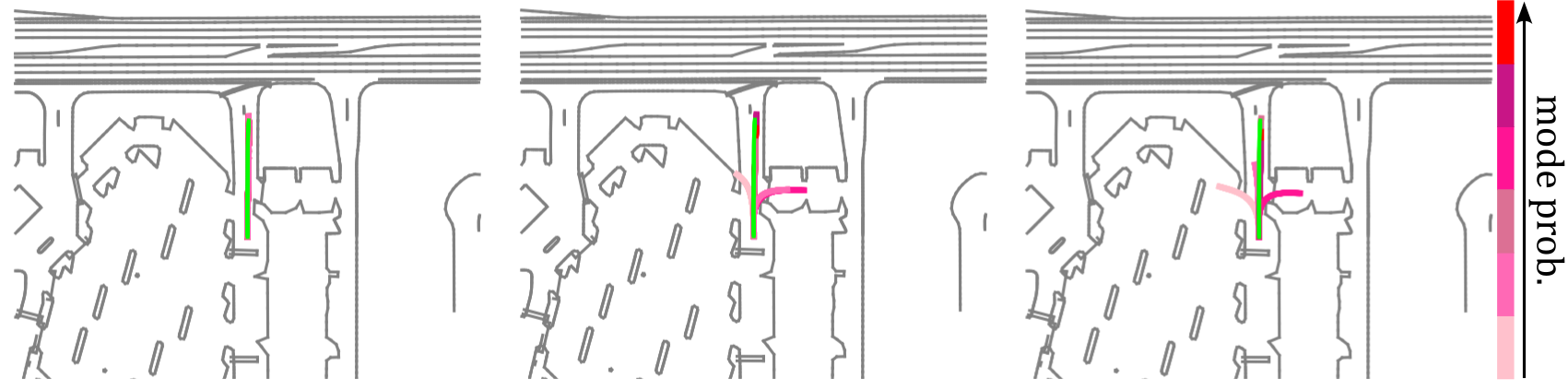}
\end{subfigure}
\vspace{-1mm}
\\
\hspace{-6mm}
\begin{minipage}[c]{0.14\linewidth}
\centering iter.=100
\end{minipage}
\begin{minipage}[c]{0.15\linewidth}
\centering iter.=1100k
\end{minipage}
\begin{minipage}[c]{0.15\linewidth}
\centering iter.=1200k
\end{minipage}
\begin{minipage}[c]{0.15\linewidth}
\centering iter.=100
\end{minipage}
\begin{minipage}[c]{0.15\linewidth}
\centering iter.=1100k
\end{minipage}
\begin{minipage}[c]{0.15\linewidth}
\centering iter.=1200k
\end{minipage}
\\
\begin{minipage}[c]{0.48\linewidth}
\centering WTA
\end{minipage}
\begin{minipage}[c]{0.48\linewidth}
\centering \textbf{\awta} (ours)
\end{minipage}
\\
\caption{
\label{plot:evolution}
\textbf{Evolution of mode distribution during training.} The predictions are obtained with Wayfomer~\cite{Wayformer} using WTA (left) or \awta (right) on Waymo Open Motion Dataset (WOMD)~\cite{wmod}. 
We observe that the effective number of hypotheses increases with the training step 
with
\awta. The ground-truth trajectories are shown in green.}
\end{figure*}

The WTA optimization procedure can be interpreted as an input-dependent gradient-based variant of the Llyod algorithm \cite{lloyd1982least, RWTA}, which is known to be sensitive to initialization and prone to mode collapse.

\subsection{Incorporating \awta to Motion Forecasting}

To alleviate sensitivity to initialization and mode collapse issues, we take inspiration from annealed MCL \cite{amcl} and incorporate \awta into motion forecasting model training.

In theory, \awta can be applied as a substitution to any WTA-based forecasting backbones. Specifically, modern motion forecasting methods~\cite{mtr,Wayformer} are mostly transformer-based. They model the forecasting heads as learnable queries followed by a decoding network with fully connected layers. The queries are initialized from either intention points, obtained from clustering the end locations of trajectories in the training datasets~\cite{mtr}, or random noise~\cite{Wayformer}. The queries interact through the attention mechanism with the map, the trajectory history, and the other agents. After that, the attended queries predict the future trajectories $f^k_{\theta}(x)$ and the associated mode probability via the decoding network.
Note that while this query-based Transformer architecture can be seen as a straightforward implementation of Multiple Choice Learning, prior work analyzing MCL \cite{lee2016stochastic, lee2017confident} only experimented with Multi-Layer Perceptrons with multiple heads.
As a result, the applicability of their findings to Transformer-based models remains empirically unverified. 

\awta softens the hard assignment between the best forecasting prediction $f^k_{\theta}(x)$ and the ground-truth $y$, described in \eqref{eq:argmin}, by using a $\mathrm{softmin}$ function, replacing the $\mathrm{min}$ operation in the selection of the Winner hypothesis. The soft assignment can be written in the form:
\begin{align}
    q_{t}\left(f^k_{\theta} \mid x, y\right) &\triangleq \frac{1}{Z_{x, y}} \exp \left(-\frac{\ell\left(f^k_{\theta}(x), y\right)}{T(t)}\right),
    \label{eq:entropy}\\
    Z_{x, y} &\triangleq \sum_{s=1}^K \exp \left(-\frac{\ell\left(f^s_{\theta}(x), y\right)}{T(t)}\right),
\end{align}
where $t$ is the training step. After the stop gradient operator being applied to $q_t$, we replace the WTA training objective ~\eqref{eq_wta} with a weighted sum of all $\ell\left(f^k_{\theta}(x), y\right)$ predicted from different heads {or, in our case, queries}: %
\begin{equation}
\mathcal{L}_t(\theta) = \sum_{k=1}^{K} q_t(f_k, x, y) \ell(f^k_{\theta}(x),y),
\label{eq:loss_expr}
\end{equation}

\textbf{Weight scheduling.}
Note that $q_t$ is scheduled by controlling the temperature $T(t)$ of the current training step $t$. Ideally, at first, each of the hypotheses should equally contribute with ${\mathrm{lim}_{t \rightarrow 0\;} q_t(f_k, x, y) = \frac{1}{K}}$: in this case, the predictions-target assignment is very \textit{soft}. In this phase, the hypotheses converge toward a conditional mean, and the \textit{effective} number of hypotheses is equal to $1$ \cite{rose1994mapping}.
As training proceeds and the temperature decreases, $\mathcal{L}_t$ converges to the standard WTA training objective, where ${\mathrm{lim}_{t \rightarrow \infty\;} q_t(f_k, x, y) = \mathbf{1}[k \in \mathrm{argmin}_s \ell(f_s(x),y)]}$ and the number of effective modes increases. Using the Boltzmann distribution for $q_t$ is justified in Proposition 2 of \cite{amcl}, as the optimal way of constraining the soft assignment to a specific level of entropy.
In practice, we observe in \autoref{plot:evolution} that the number of modes effectively increases during training. This is because the hypotheses better cover the space of possible future trajectories.

In practice, to allow $q_t$ to slowly converge to the loss of WTA at the end of training, we control the temperature $T(t)$ {using} an exponential temperature scheduler: 
\begin{align}
    T(t) = T_{0}\rho^{t},
    \label{eq:exp_scheduler}
\end{align}
where $t$ is the current number of training steps. The initial temperature $T_{0}$, and the speed of temperature decay $\rho$ are carefully ablated in the following section.
Remarkably, as \awta helps gradually distribute the predictions to cover all plausible trajectories, we significantly reduce the number of training hypotheses (i.e., number of queries, from 64 to 6) and discard the selection process, yielding a better forecasting performance with fewer training queries. 
Experimental proofs and ablation studies are given in~\autoref{sec:exp}.

\section{Experimental Results}
\label{sec:exp}
In this section, we first describe the experimental protocol and the training details in~\autoref{subsec:evaluation_protocol}.
We then demonstrate the benefits of \awta in motion forecasting in~\autoref{subsec:useful_properties}. Then in~\autoref{subsec:baseline_ablation}, we compare \awta with existing WTA variants, showcasing its superior performance for recent motion forecasting models in large-scale datasets, followed by an ablation study of important \awta components.

\begin{table*}[t]
\setlength\extrarowheight{-3pt}
\centering
\caption{\textbf{Comparison between \awta and the standard WTA.} The use of \awta, over the default WTA loss, consistently improves MTR \cite{mtr} and Wayformer \cite{Wayformer} over Argoverse 2 and WOMD, across all metrics. Results 
on the validation sets.}
\begin{NiceTabular}{l@{\hskip 4pt}cccccccc}
\toprule 
& \multicolumn{4}{c}{\textbf{Argoverse 2} \cite{argoverse2}} & \multicolumn{4}{c}{\textbf{WOMD} \cite{wmod}}   \\
\cmidrule(lr){2-5}\cmidrule(lr){6-9}
 & Brier-FDE$\downarrow$ & minADE$\downarrow$  & minFDE$\downarrow$ & MissRate$\downarrow$ & Brier-FDE$\downarrow$ & minADE$\downarrow$ & minFDE$\downarrow$ & MissRate$\downarrow$  \\

\midrule

MTR \cite{mtr} $w/$ WTA (default) & 2.12 & 0.85 & 1.68 &0.30 & 2.19 & 0.76 & 1.74 & 0.31  \\ %

\rowcolor{buscasota}
\makecell{MTR \cite{mtr} \plusours{} \textbf{(ours)} \vspace*{-1mm}\\ } & 
\makecell{\textcolor{blue}{\textbf{2.08}}\\ \gooddelta{\text{- }1.89\%}} &
\makecell{\textcolor{blue}{\textbf{0.77}}\\ \gooddelta{\text{- }9.41\%}} &
\makecell{\textcolor{blue}{\textbf{1.46}}\\ \gooddelta{\text{- }13.10\%}} &
\makecell{\textcolor{blue}{\textbf{0.19}}\\ \gooddelta{\text{- }36.67\%}} &
\makecell{\textcolor{blue}{\textbf{1.98}}\\ \gooddelta{\text{- }9.59\%}} &
\makecell{\textcolor{blue}{\textbf{0.63}}\\ \gooddelta{\text{- }17.11\%}} &
\makecell{\textcolor{blue}{\textbf{1.34}}\\ \gooddelta{\text{- }22.99\%}} &
\makecell{\textcolor{blue}{\textbf{0.18}}\\ \gooddelta{\text{- }41.94\%}} \\

\midrule

Wayformer \cite{Wayformer} $w/$ WTA (default) & 2.19 & 0.79 & 1.57 & 0.24 & 2.10 & 0.66 & 1.46 & 0.22  \\
 \rowcolor{buscasota}
\makecell{Wayformer \cite{Wayformer} \plusours{} \textbf{(ours)} \vspace*{-1mm}\\ } & 
\makecell{\textcolor{blue}{\textbf{2.16}}\\ \gooddelta{\text{- }1.37\%}} & 
\makecell{\textcolor{blue}{\textbf{0.78}}\\ \gooddelta{\text{- }1.27\%}} & 
\makecell{\textcolor{blue}{\textbf{1.53}}\\ \gooddelta{\text{- }2.55\%}} & 
\makecell{\textcolor{blue}{\textbf{0.22}}\\ \gooddelta{\text{- }8.33\%}} & 
\makecell{\textcolor{blue}{\textbf{2.09}}\\ \gooddelta{\text{- }0.48\%}} & 
\makecell{\textcolor{blue}{\textbf{0.65}}\\ \gooddelta{\text{- }1.52\%}} & 
\makecell{\textcolor{blue}{\textbf{1.45}}\\ \gooddelta{\text{- }0.68\%}} & 
\makecell{\textcolor{blue}{\textbf{0.21}}\\ \gooddelta{\text{- }4.55\%}} \\
\bottomrule
\end{NiceTabular}
\vspace{-3mm}
\label{tab:motion_forecasting_awta}
\end{table*}

\subsection{Experimental protocol and implementation details} 
\label{subsec:evaluation_protocol}
\noindent \textbf{Evaluation protocol.} For fair comparisons, we follow the standardized evaluation protocol proposed in UniTraj \cite{feng2024unitraj}.
Models are tasked to predict future trajectories (up to 6 trajectories)
for the next 6 seconds at 10Hz (i.e., $L=60$), based on the scene (e.g., HD maps) and agent trajectories from the past 2 seconds.
We use the following metrics: $\bullet$ $\textbf{minADE}_k$, minimum over $k=6$ predictions of the Average Distance Error (the average of point-wise L$^2$ distances between the prediction and ground-truth forecasts), $\bullet$ $\textbf{minFDE}_k$, minimum over $k=6$ predictions of the Final Distance Error, $\bullet$ MissRate $\textbf{MR}_{k@x}$, ratio of forecasts with $\text{minFDE}_k>2$ meters, and lastly, $\bullet$ $\textbf{brier-FDE}$, sum of $\text{minFDE}_k$ and $(1-\delta)^2$, with $\delta$ the hypothesis score. Performance is reported on the validation sets.

\noindent\textbf{Models and datasets.}
{We integrate \awta into two recent motion forecasting models: MTR~\cite{mtr} (60M parameters) and Wayformer~\cite{Wayformer} (16M parameters), provided in UniTraj \cite{feng2024unitraj}. Vanilla MTR is trained with 64 hypotheses and selects 6 of them with NMS, while Vanilla Wayformer uses only 6 hypotheses for training.
We replace all WTA objectives (for both final and intermediate predictions) in the original implementation by \awta with an exponential temperature scheduler, and the number of hypotheses is fixed to 6 for training and inference.}
We note that the hypothesis score loss (i.e., cross-entropy loss) remains unchanged. 
We conduct experiments on two large-scale real-world datasets: Argoverse 2 \cite{argoverse2}, containing 180k trajectories, and Waymo Open Motion Dataset (WOMD) \cite{wmod}, with 1.8M trajectories.\\
\textbf{Training details.}
All models are trained until their convergence based on $\textbf{minFDE}_k$ (around 100 epochs).
Training time with \awta is $\sim$4 minutes (MTR) or 3 minutes (Wayformer) per epoch for Argoverse 2 and  49 minutes (MTR) or 29 minutes (Wayformer) for WOMD, on a node of 8x A100 GPUs. The total batch size is fixed to 256 for MTR and 128 for Wayformer, as in \cite{feng2024unitraj}.
We perform a grid search on the initial temperature and temperature decay, selecting the best-performed hyper-parameters, as discussed in~\autoref{subsec:baseline_ablation}.
The forecasting performance is not highly sensitive to these hyperparameters within a reasonable range.
For WTA variants, we follow the descriptions in their respective papers and reimplement them to MTR \cite{mtr} and Wayformer \cite{Wayformer}. %

\subsection{Main results: benefits of \awta.}
\label{subsec:useful_properties} 

\textbf{Using \awta boosts forecasting performances.}
We present the performance of MTR~\cite{mtr} and Wayformer~\cite{Wayformer}, with the default WTA loss and our \awta proposal, on Argoverse 2~\cite{argoverse2} and WOMD~\cite{wmod} in \autoref{tab:motion_forecasting_awta}.
The first conclusion is that the use of \awta, in place of standard WTA, systematically improves both methods, on the two datasets, across all metrics.
For instance, using \awta yields a decrease of 23\% in minFDE for MTR on WOMD, and a decrease of 8.3\% in MissRate for Wayformer on Argoverse 2.
This benefit is consistently observed and particularly visible for MTR, as {this may be explained by the bigger model size and auxiliary losses in all intermediate layers of MTR.}
\begin{figure}[t]
    \centering
    \includegraphics[width=0.9\linewidth]{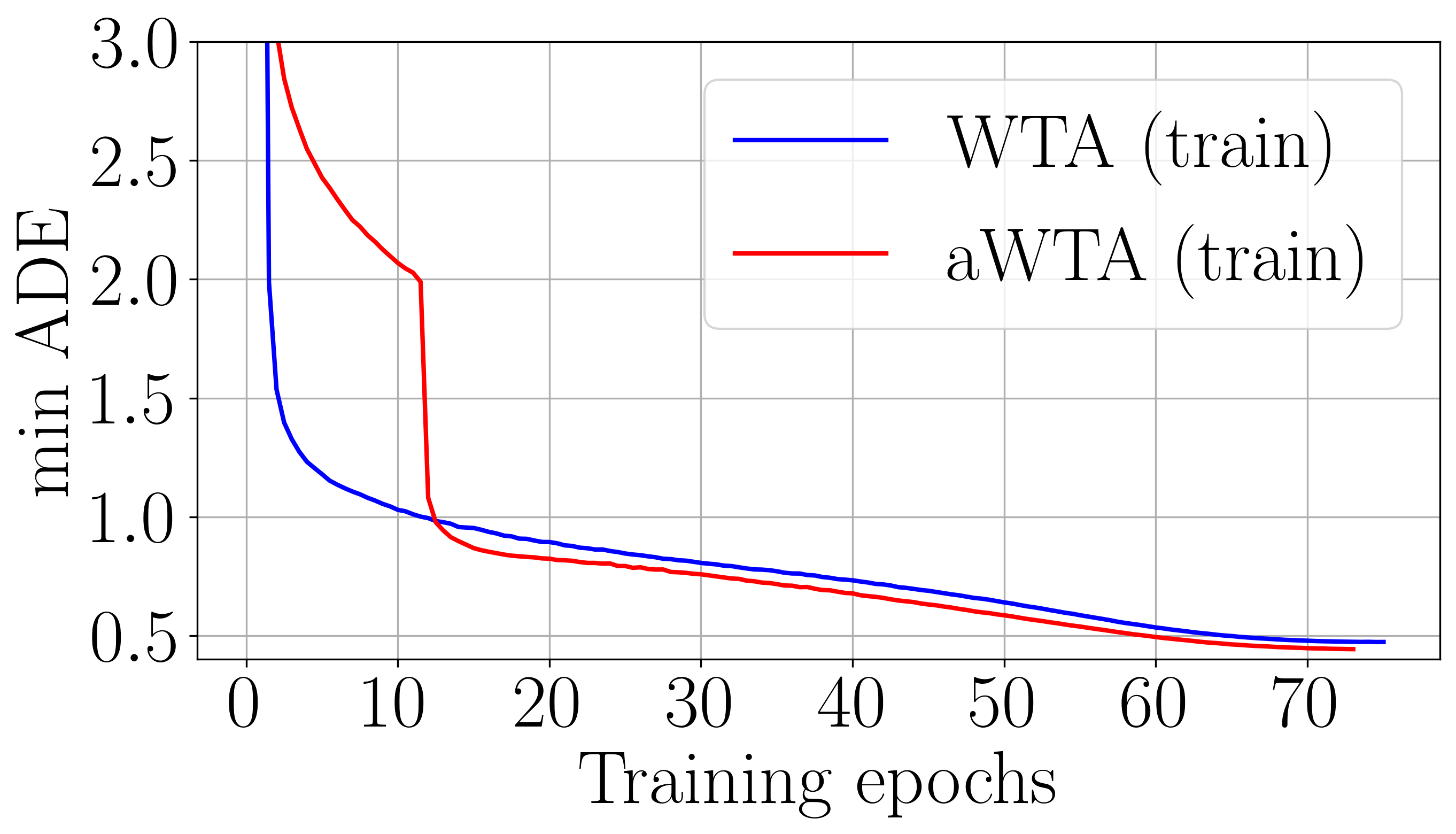}
    \caption{
    \textbf{Phase transition with aWTA Loss.} 
    Evolution of (averaged) minADE during training of Wayformer \cite{Wayformer} on Argoverse 2 \cite{argoverse2}, comparing WTA (blue) and aWTA (red) training setups. A sudden drop in error is observed around epoch $12$, consistent with the expected behavior of the deterministic annealing procedure. Here, we see that the aWTA converges to a better training fit compared to 
    WTA.}\label{fig:training_dynamics}
\end{figure}

\textbf{Phase transitions.}
As the temperature decreases, we observe abrupt drops in the loss value, as shown in \autoref{fig:training_dynamics}.
These drops can be theoretically explained as `phase transitions' \cite{rose1990deterministic, rose1994mapping, amcl}.
At certain critical temperatures, sudden 
improvement occurs in the metrics.
{These phase transitions correspond to moments when the effective number of hypotheses changes.}
Our observations align with the theoretical insights from \cite{amcl}.

\begin{table}[t]
\setlength\extrarowheight{-3pt}
\centering
\caption{\textbf{\awta: better performance, less hypotheses.} We compare training with \awta and 6 hypotheses to training with WTA with 6 hypotheses or 64 hypotheses after `Non-Maximum Suppression' post-processing (`+ NMS'). The results are reported in the validation sets of Argoverse 2~\cite{argoverse2} and WOMD~\cite{wmod} using MTR~\cite{mtr}. 'time' indicates the training time per epoch in minutes.}
\resizebox{0.98\linewidth}{!}{%
\begin{NiceTabular}{l@{\hskip 6pt}c@{\hskip 4pt}c@{\hskip 4pt}c@{\hskip 4pt}c@{\hskip 4pt}c@{\hskip 4pt}c@{\hskip 4pt}c@{\hskip 4pt}}
\toprule
& & \multicolumn{2}{c}{\textbf{\#hypo.}} &  \multicolumn{4}{c}{\textbf{Argoverse 2} \cite{argoverse2}}  \\

\textbf{MTR} $w/$ & time (min.) & train & infer. & Brier-FDE$\downarrow$ & minADE$\downarrow$ &minFDE$\downarrow$ & MissRate$\downarrow$   \\
\cmidrule(lr){5-8}

  WTA & 3 & 6 & 6 & 3.34 &  1.36  &	3.08 &  0.56  \\
  WTA + NMS & 5 & 64 & 6  & 2.12 & 0.85 & 1.68 & 0.30  \\
\rowcolor{buscasota}
\makecell{\textbf{\awta (ours)}\\ } & 4 & 6 & 6 &
\makecell{\textcolor{blue}{\textbf{2.08}}} & 
\makecell{\textcolor{blue}{\textbf{0.77}}} &
\makecell{\textcolor{blue}{\textbf{1.46}}} &
\makecell{\textcolor{blue}{\textbf{0.19}}} \\

\midrule

& &\multicolumn{2}{c}{\textbf{\#hypo.}} &  \multicolumn{4}{c}{\textbf{WOMD} \cite{wmod}}  \\

\textbf{MTR} $w/$ & time (min.) &  train & infer. & Brier-FDE$\downarrow$ & minADE$\downarrow$ & minFDE$\downarrow$ & MissRate$\downarrow$   \\
\cmidrule(lr){5-8}

WTA & 41 &6 & 6 & 3.40 & 1.27  & 3.22   & 0.58   \\
WTA + NMS &61 &64 & 6 & 2.19 & 0.76 & 1.74 & 0.31  \\

\rowcolor{buscasota}
\makecell{\textbf{\awta (ours)}\\ } &49 & 6 & 6 &
\makecell{\textcolor{blue}{\textbf{1.98}}} &
\makecell{\textcolor{blue}{\textbf{0.63}}} &
\makecell{\textcolor{blue}{\textbf{1.34}}} & 
\makecell{\textcolor{blue}{\textbf{0.18}}} \\

\bottomrule
\end{NiceTabular}}
\label{tab:awta_better_loss}
\end{table}

\textbf{Discussion on the number of hypotheses.}
Motion forecasting methods trained with WTA often suffer from mode collapse when trained with a small number of hypotheses, e.g., 6, as observed in \autoref{fig:teaser} and quantified with the poor results in \autoref{tab:awta_better_loss} (6 hypotheses for both training and inference).
To mitigate this issue, methods typically train with more hypotheses, e.g., 64, and then use post-processing techniques, such as Non-Maximum Suppression (NMS) or clustering, to select the top-$K$ best hypotheses (in our case, $K=6$ following the experimental protocol).
However, this approach requires careful design and tuning of the post-processing step.
In contrast, with \awta, we use the same number of hypotheses during training and inference, $K=6$. This approach yields better performance and save $\sim$20\% the training time than methods that artificially increase the number of hypotheses during training and that require post-processing of predictions, as seen in~\autoref{tab:awta_better_loss}.

\begin{table}[h]
\setlength\extrarowheight{-3pt}
\centering
\caption{\textbf{Comparison between \awta, WTA variants}, namely RWTA~\cite{RWTA}, EWTA~\cite{EWTA}, and DAC~\cite{dac}, and Linear \awta. Results are reported on the validation sets of Argoverse 2~\cite{argoverse2} and WOMD~\cite{wmod} using MTR~\cite{mtr} and Wayformer~\cite{Wayformer}.}
\resizebox{0.98\linewidth}{!}{%
\begin{NiceTabular}{l@{\hskip 6pt}cccc}
\toprule

& \multicolumn{4}{c}{\textbf{Argoverse 2} \cite{argoverse2}}  \\
\textbf{MTR} \cite{mtr} & Brier-FDE$\downarrow$ & minADE$\downarrow$ & minFDE$\downarrow$ & MissRate$\downarrow$   \\
\cmidrule(lr){2-5}

$w/$  WTA & 2.12 & 0.85 & 1.68 & 0.30  \\
$w/$  RWTA~\cite{RWTA} &  2.30 & 0.83 & 1.63 & 0.20\\
$w/$  EWTA~\cite{EWTA} & 2.09& \textbf{0.77} & 1.49 &0.20  \\
$w/$  DAC~\cite{dac} &   2.10 & 0.78 & 1.50 & 0.20   \\
$w/$  Linear \awta & 4.44& 1.58& 3.75& 0.52\\
\rowcolor{buscasota}
\makecell{\plusours{} \textbf{(ours)}\\ } & 
\makecell{\textcolor{blue}{\textbf{2.08}}} &
\makecell{\textcolor{blue}{\textbf{0.77}}} &
\makecell{\textcolor{blue}{\textbf{1.46}}} &
\makecell{\textcolor{blue}{\textbf{0.19}}} \\

& \multicolumn{4}{c}{\textbf{WOMD} \cite{wmod}}  \\
\textbf{MTR} \cite{mtr} & Brier-FDE$\downarrow$ & minADE$\downarrow$ & minFDE$\downarrow$ &MissRate$\downarrow$   \\
\cmidrule(lr){2-5}

$w/$  WTA & 2.19 & 0.76 & 1.74 & 0.31  \\
$w/$  RWTA~\cite{RWTA} &  2.17 & 0.70 & 1.49 & 0.19 \\
$w/$  EWTA~\cite{EWTA} & 2.00 & 0.65 & 1.37 &0.19  \\
$w/$  DAC~\cite{dac} & 2.15 & 0.71 & 1.57 & 0.25  \\
$w/$  Linear \awta &  2.15&0.71&1.57&0.25 \\
\rowcolor{buscasota}
\makecell{\plusours{} \textbf{(ours)}\\ } & 
\makecell{\textcolor{blue}{\textbf{1.98}}} & 
\makecell{\textcolor{blue}{\textbf{0.63}}} &
\makecell{\textcolor{blue}{\textbf{1.34}}} &
\makecell{\textcolor{blue}{\textbf{0.18}}} \\
\midrule

&  \multicolumn{4}{c}{\textbf{Argoverse 2} \cite{argoverse2}}\\
\textbf{Wayformer} \cite{Wayformer} & Brier-FDE$\downarrow$ & minADE$\downarrow$ & minFDE$\downarrow$ & MissRate$\downarrow$   \\
\cmidrule(lr){2-5}
$w/$  WTA  & 2.19 & 0.79 & 1.57 & 0.24 \\ 
$w/$  RWTA~\cite{RWTA} &  2.30 & 0.81& 1.64& 0.25 \\
$w/$  EWTA~\cite{EWTA} & 2.20 &0.79 &1.58 &0.24  \\
$w/$  DAC~\cite{dac} &   2.18& 0.78& 1.56& 0.24 \\
$w/$  Linear \awta & 3.43 & 1.31& 2.73& 0.35  \\
\rowcolor{buscasota}
\makecell{\plusours{} \textbf{(ours)} \\ } & 
\makecell{\textcolor{blue}{\textbf{2.16}} } &
\makecell{\textcolor{blue}{\textbf{0.78}} } &
\makecell{\textcolor{blue}{\textbf{1.53}} } &
\makecell{\textcolor{blue}{\textbf{0.22}}}  \\

&   \multicolumn{4}{c}{\textbf{WOMD} \cite{wmod}}\\
\textbf{Wayformer} \cite{Wayformer} & Brier-FDE$\downarrow$ & minADE$\downarrow$ & minFDE$\downarrow$ & MissRate$\downarrow$   \\
\cmidrule(lr){2-5}
$w/$  WTA  & 2.10 & 0.66 & 1.46 & 0.22 \\ 
$w/$  RWTA~\cite{RWTA} & 2.23&0.69&1.55& \textbf{0.21} \\
$w/$  EWTA~\cite{EWTA} &2.12&0.67&1.50&0.23 \\
$w/$  DAC~\cite{dac} &2.16&0.68&1.53&0.24 \\
$w/$  Linear \awta & 5.25&1.66&4.55&0.66 \\
\rowcolor{buscasota}
\makecell{\plusours{} \textbf{(ours)} \\ } & 
\makecell{\textcolor{blue}{\textbf{2.09}} } &
\makecell{\textcolor{blue}{\textbf{0.65}} } &
\makecell{\textcolor{blue}{\textbf{1.45}} } &
\makecell{\textcolor{blue}{\textbf{0.21}}}  \\

\bottomrule
\end{NiceTabular}}
\vspace{-4mm}
\label{tab:baselines}
\end{table}

\subsection{Baselines and ablation.}
\label{subsec:baseline_ablation}

\noindent \paragraph{Comparison to other WTA variants}
\label{subsec:baseline_sota}

Although some WTA variants have been proposed to alleviate the instability of WTA, it is worth noting that most of the WTA variants (RWTA \cite{RWTA}, EWTA \cite{EWTA} and DAC \cite{dac}) have primarily been tested in simulated datasets with small models, e.g., ResNet-18, and have rarely been tested on real-world datasets. To our knowledge, we conducted the first comparison of different WTA variants on large-scale real-world datasets.

From \autoref{tab:baselines}, we observe that our proposed \awta consistently outperforms other variants for MTR and Wayformer on Argoverse 2 and WOMD across all metrics.
In particular, we find that no other WTA variant improves Wayformer, especially in WOMD, while the use of the \awta loss does.
This might be because these methods are more sensitive to the hypothesis initialization, which is random noise in Wayformer. Moreover, aWTA is unique in having an infinite, uncountable set $\mathcal{V}$ for $q_t$ values, providing more flexibility.
\begin{figure*}[t]
\centering
\begin{minipage}[c]{0.09\linewidth}
\centering 
\end{minipage}
\begin{minipage}[c]{0.192\linewidth}
\centering WTA
\end{minipage}
\hspace{-0.2cm}
\begin{minipage}[c]{0.192\linewidth}
\centering RWTA \cite{RWTA}
\end{minipage}
\hspace{-0.3cm}
\begin{minipage}[c]{0.192\linewidth}
\centering EWTA \cite{EWTA}
\end{minipage}
\hspace{-0.6cm}
\begin{minipage}[c]{0.192\linewidth}
\centering DAC \cite{dac}
\end{minipage}
\hspace{-0.5cm}
\begin{minipage}[c]{0.192\linewidth}
\centering \textbf{\awta} (ours)
\end{minipage}

\begin{subfigure}{\linewidth}
\begin{minipage}[c]{0.09\linewidth}
\centering MTR\\Argoverse 2
\end{minipage}
\begin{minipage}[c]{0.881\linewidth}
\includegraphics[width=\textwidth]{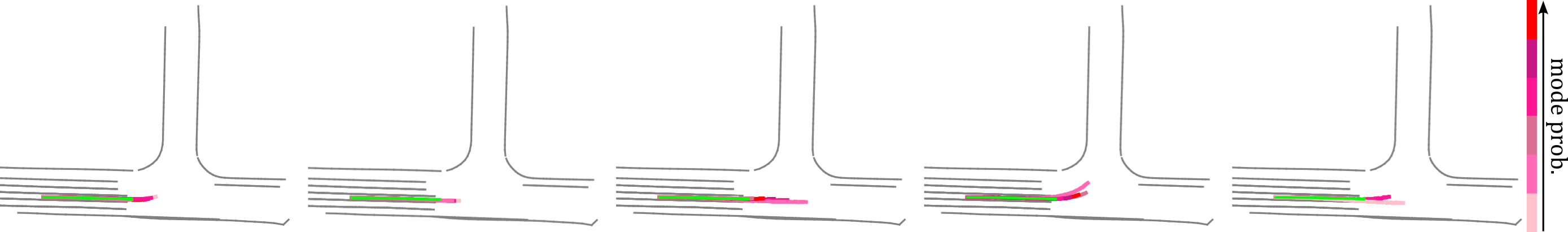}
\end{minipage}
\end{subfigure}
\\
\begin{subfigure}{\linewidth}
\begin{minipage}[c]{0.09\linewidth}
\centering Wayformer\\Argoverse 2
\end{minipage}
\begin{minipage}[c]{0.881\linewidth}
\includegraphics[width=\textwidth]{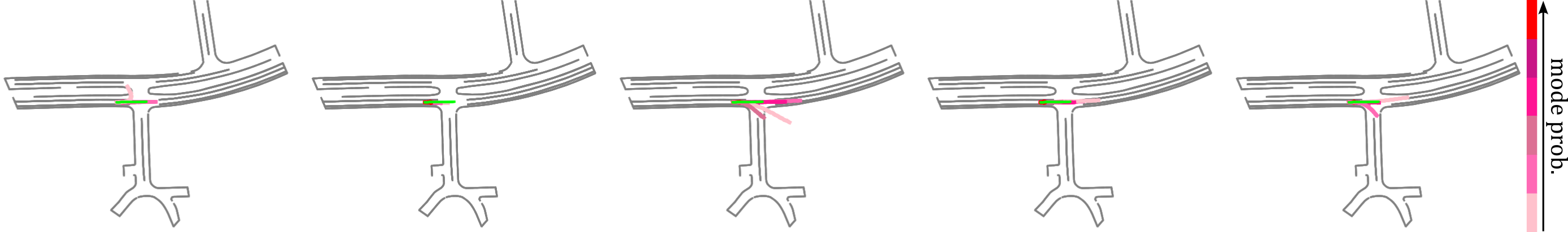}
\end{minipage}
\end{subfigure}
\\
\begin{subfigure}{\linewidth}
\begin{minipage}[c]{0.09\linewidth}
\centering MTR\\WOMD
\end{minipage}
\begin{minipage}[c]{0.881\linewidth}
\includegraphics[width=\textwidth]{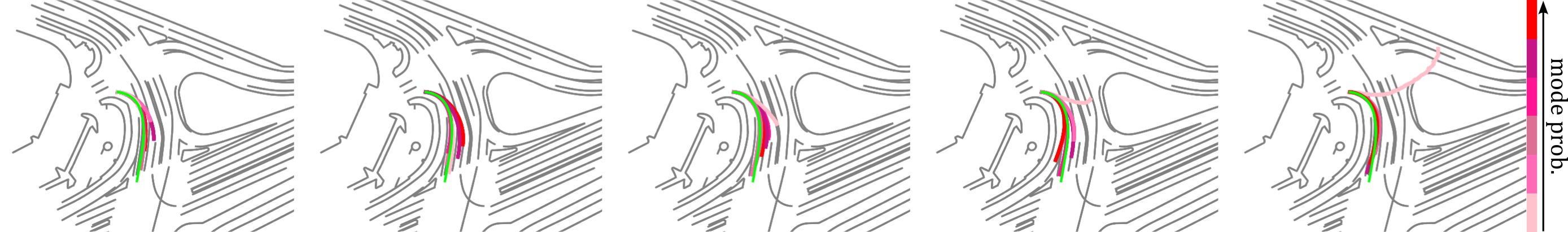} 
\end{minipage}
\end{subfigure}
\\
\begin{subfigure}{\linewidth}
\begin{minipage}[c]{0.09\linewidth}
\centering Wayformer\\WOMD
\end{minipage}
\begin{minipage}[c]{0.881\linewidth}
\includegraphics[trim={0 1cm 0 0},clip,width=\textwidth]{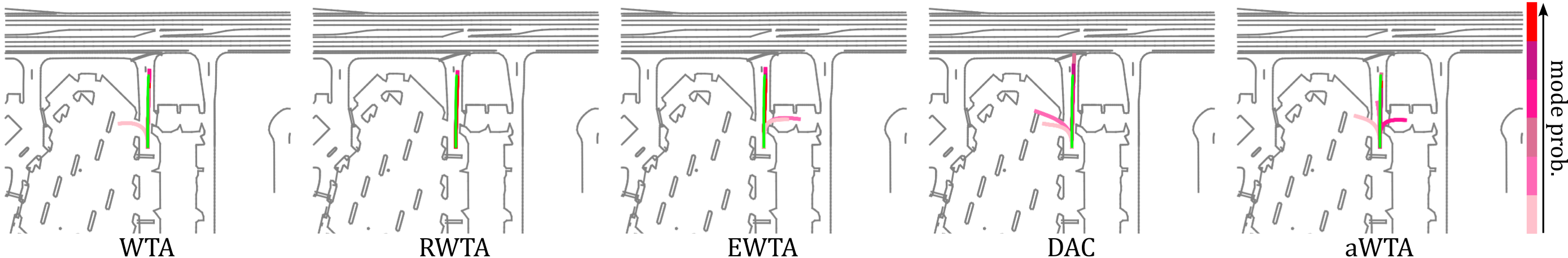}
\end{minipage}
\end{subfigure}
\caption{\textbf{Qualitative comparison between \awta and WTA variants.} Predictions are shown on Argoverse 2 \cite{argoverse2} (rows 1 and 2) and WOMD \cite{wmod} (rows 3 and 4) for MTR \cite{mtr} (rows 1 and 3) and Wayformer \cite{Wayformer} (rows 2 and 4) models. The ground-truth trajectories are shown in green.
}
\label{fig:qualitative_results}
\vspace{-2mm}
\end{figure*}

Additionally, we present qualitative examples of predictions given by MTR trained with different WTA variants in \autoref{fig:qualitative_results}.
We observe that MTR with \awta tends to cover a wider range of feasible trajectories compared to other WTA variants.
In contrast, other variants distribute less widely over different plausible trajectories and struggle to cover the diverse ground-truth trajectories.

\noindent\paragraph{Ablation study}
\awta employs a temperature scheduling scheme where the temperature gradually decreases during training.
Eventually, this scheme converges to an optimization process equivalent to the standard WTA approach. This training scheme introduces two key parameters: (1) the temperature scheduler which determines how quickly the temperature decreases and the pattern of this decrease, and (2) the initial temperature value. We discuss below the choice and impact of these parameters on the performance.

\noindent\textbf{Temperature scheduler.} To implement the \awta loss, we ablate on two types of scheduler, i.e., exponential \eqref{eq:exp_scheduler} or linear \eqref{eq:linear_scheduler} schedulers:
\begin{equation}
\vspace{-1mm}
T(t) = T_{0}\left(1 - \frac{t}{100}\right) \mathbf{1}(t < 100) + T_f \mathbf{1}(t \geq 100),
\label{eq:linear_scheduler}
\end{equation}
where $T_f = 1e-8$.
Both schedulers start from the same initial temperature $T_{0}$ (10 for MTR and 8 for Wayformer) and $\rho$ is set to 0.834 and 0.89 for MTR and Wayformer respectively.
\begin{figure}[h]
\centering
\begin{subfigure}{0.49\linewidth} 
\includegraphics[trim={0cm 0cm 0 0.0cm},clip, width=\linewidth]{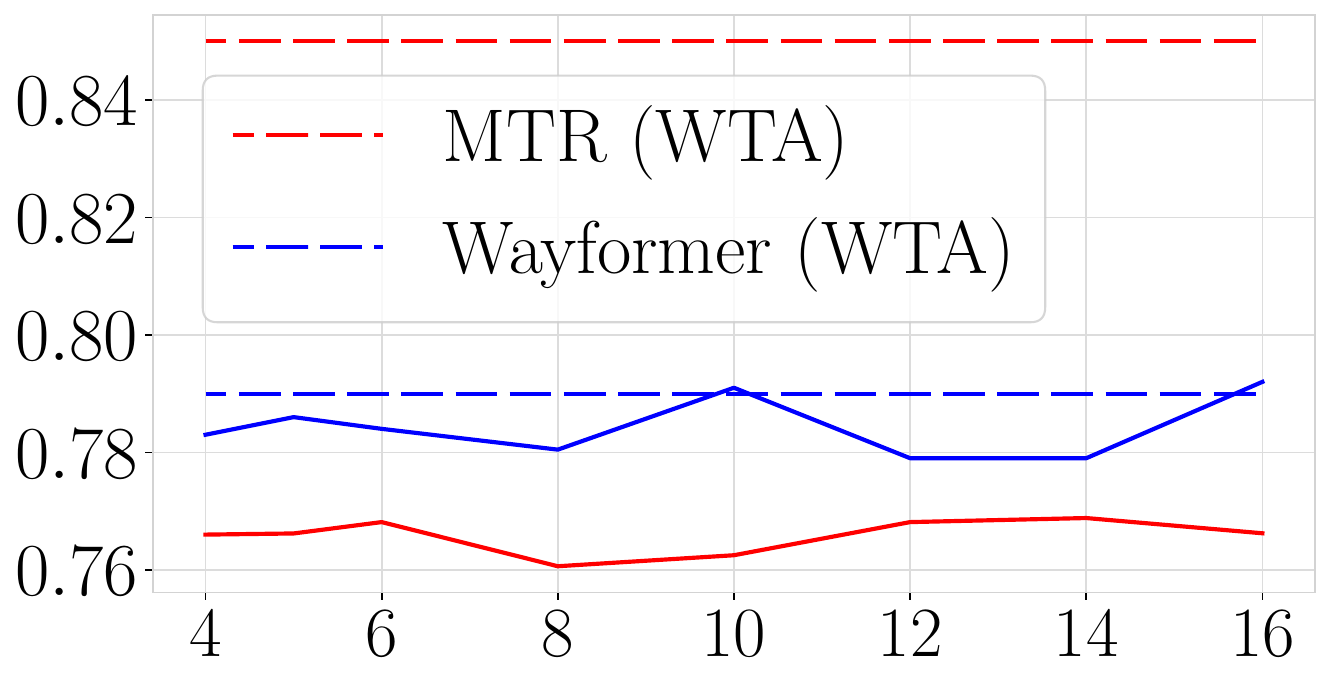}
    \vspace{-0.6cm} 
  \caption{minADE $\downarrow$}
\end{subfigure}
\centering
\begin{subfigure}{0.49\linewidth}
\includegraphics[trim={0cm 0cm 0 0cm},clip, width=\linewidth]{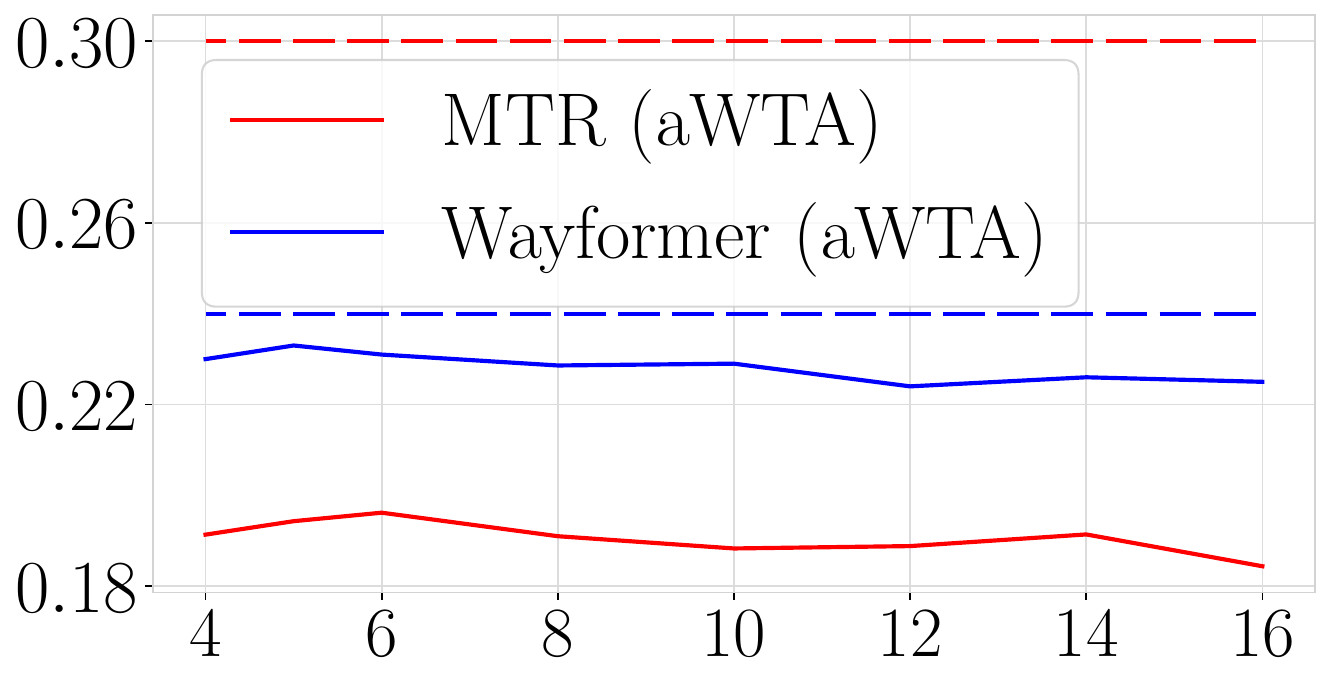}
     \vspace{-0.6cm} 
  \caption{MissRate $\downarrow$}
\end{subfigure}
\caption{\textbf{Impact of the initial temperature value $T_0$ ($x$ axis) on the performance ($y$ axis)} for \awta in two different methods, MTR \cite{mtr} (in red) and Wayformer \cite{Wayformer} (in blue). Results are obtained with Argoverse 2 validation set~\cite{argoverse2}. The dotted lines are baselines trained with default WTA loss. %
}
\label{fig:ablation_init_temp}
\vspace{-2mm}
\end{figure}

\begin{figure}[h]
\centering
\begin{subfigure}{0.49\linewidth} 
\includegraphics[trim={0cm 0cm 0 0cm},clip, width=\linewidth]{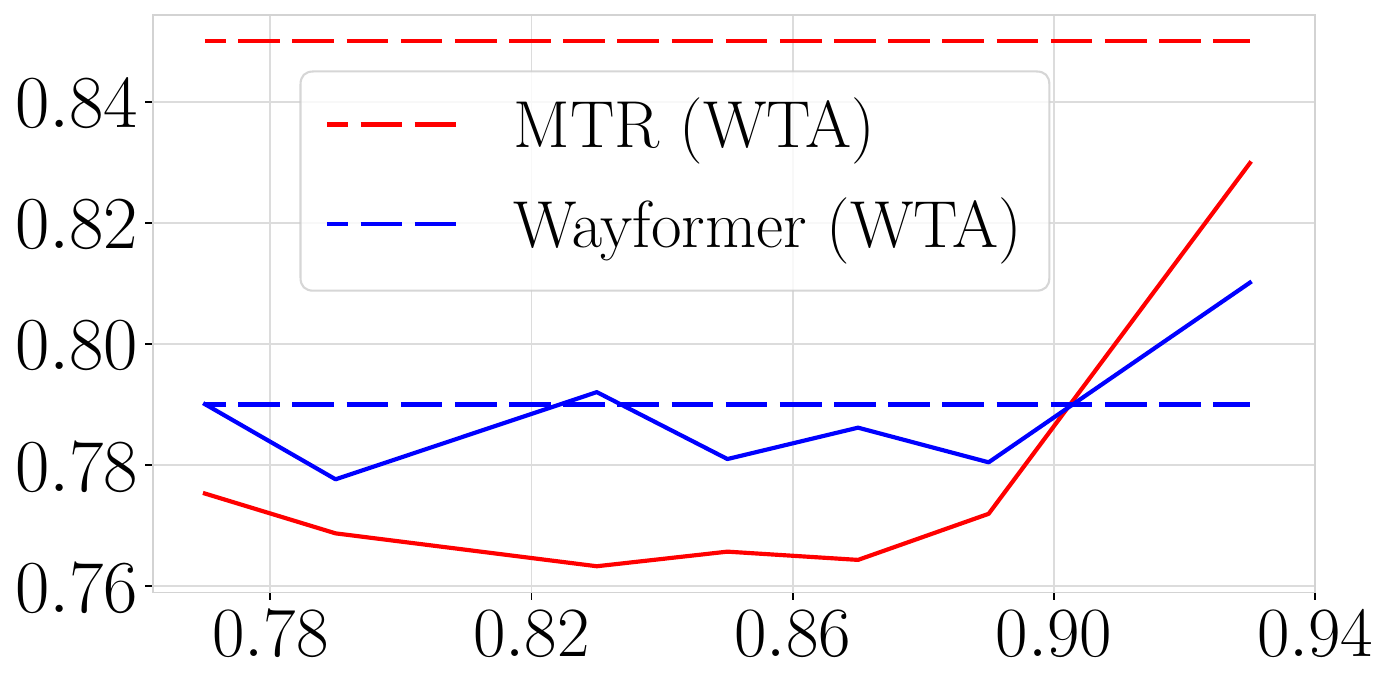}
  \caption{minADE $\downarrow$}
\end{subfigure}
\centering
\begin{subfigure}{0.49\linewidth}
\includegraphics[trim={0cm 0cm 0 0cm},clip, width=\linewidth]{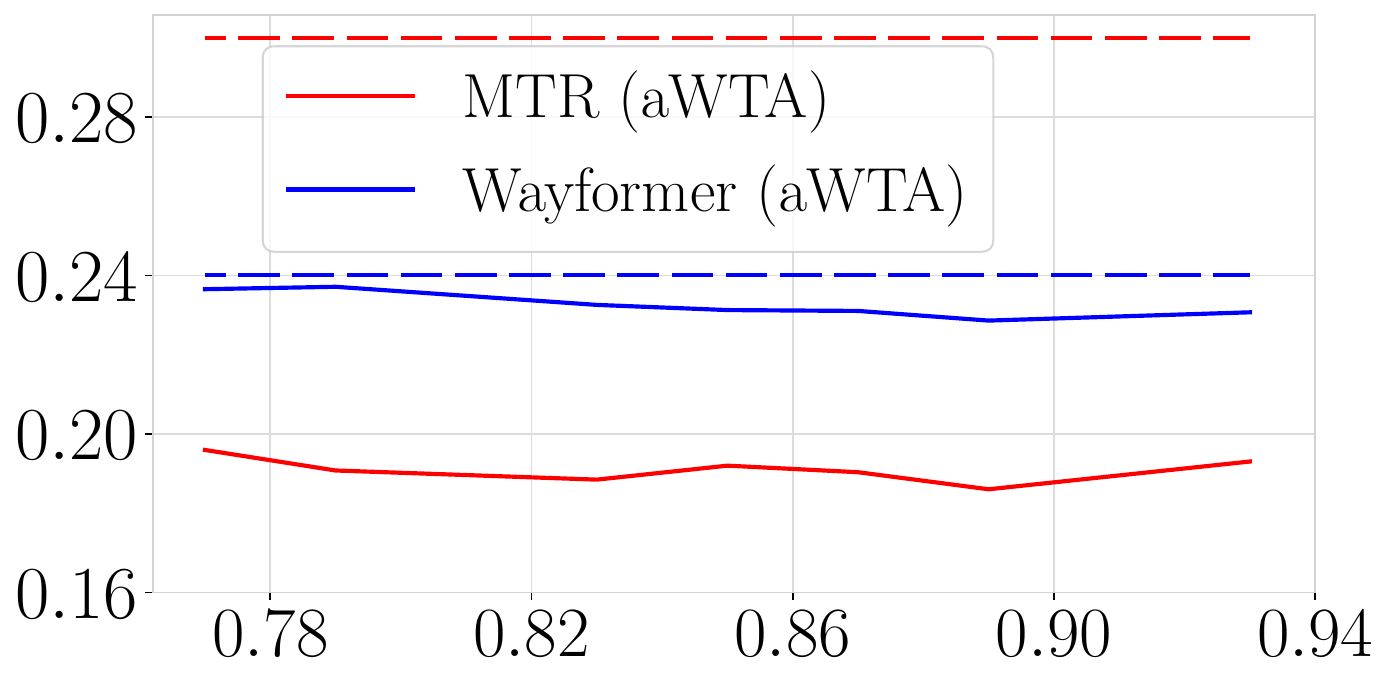}
  \caption{MissRate $\downarrow$}
\end{subfigure}
\caption{\textbf{Impact of temperature decay rate $\rho$ ($x$ axis) on the performance ($y$ axis)} for \awta with an exponential scheduler in two different methods, MTR \cite{mtr} (in red) and Wayformer \cite{Wayformer} (in blue). Results are on Argoverse 2 valset~\cite{argoverse2}. The dotted lines are baselines with WTA loss.}
\label{fig:ablation_temp_decay}
\vspace{-5mm}
\end{figure}

From the results shown in \autoref{tab:baselines} (Linear \awta vs.\ \awta), we observe that the exponential scheduler, which provides a slower slope in the end, achieves much better empirical performance. Therefore, we choose the exponential temperature scheduler for our implementation.

\noindent\textbf{Initial temperature.} 
We study the sensitivity of the forecasting performance to the initial temperature of \awta.
For these experiments, we use the exponential scheduler with a temperature decay fixed at 0.834 for both MTR and Wayformer.
From the results shown in \autoref{fig:ablation_init_temp}, we observe that the performance gains brought by \awta are maintained across a wide range of initial temperature values, showing that \awta is relatively insensitive to this hyperparameter.

\noindent\textbf{Temperature decay.} 
Similarly, with a fixed initial temperature ($10$ for MTR and $8$ for Wayformer) we study the performance impact of the rate of the temperature decay, with exponential scheduling.
The results are shown in \autoref{fig:ablation_temp_decay}. 

From \autoref{fig:ablation_init_temp}, \autoref{fig:ablation_temp_decay} and our empirical experience, we observe that \awta improves over the baselines {(dotted lines of the model trained with WTA), as long as the initial temperature is within the order of magnitude, as the average distance between the predictions and the ground truth, and the temperature decay is around $0.8$.}
This demonstrates that \awta is not too sensitive to such hyperparameters, facilitating its adoption in different motion forecasting methods.

\section{Conclusion}
In the past years, the motion forecasting literature has widely and overwhelmingly adopted the WTA training loss.
In fact, this training scheme is both simple and {quite} effective in capturing the many modes of the output, even when the modes have unbalanced probabilities.
However, WTA comes with well-established training instabilities that have been largely ignored in the community. Instead, we collectively adopted cumbersome workarounds that increase the computation cost and make using the predictions more difficult.
In this paper, we examine the root issues with WTA for motion forecasting and propose to tackle them directly using a better, while simple, training scheme.
In doing so, we show that we do not need an additional number of hypotheses, nor the post-selection step, to attain competitive performance.
Our work highlights the importance for the community to further investigate and improve upon WTA training, specifically in the context of motion forecasting, as we believe it to be currently a bottleneck for motion forecasting methods.
\section*{ACKNOWLEDGMENT}
This work was supported by the ANR MultiTrans (ANR-21-CE23-0032) and CINES (HORS DARI N°A0141014181). This research received the support of EXA4MIND project, funded by a European Union's Horizon Europe Research and Innovation Programme, under Grant Agreement N°101092944. Views and opinions expressed are however those of the author(s) only and do not necessarily reflect those of the European Union or the European Commission. Neither the European Union nor the granting authority can be held responsible for them.
\newpage
{\small
\bibliographystyle{IEEEtran}
\bibliography{biblio}
}
\end{document}